\documentclass[10pt,twocolumn,letterpaper]{article}

\usepackage{wacv}
\usepackage{times}
\usepackage{epsfig}
\usepackage{graphicx}
\usepackage{amsmath}
\usepackage{amssymb}
\usepackage{booktabs}
\usepackage{tabularx}
\usepackage{multirow,bigdelim}
\usepackage[font=footnotesize,labelfont=bf]{caption}

\usepackage{xcolor,colortbl}
\usepackage{pifont}
\usepackage{float}

\newcommand{\framework}{
\begin{figure*}[t]
  \centering
  \includegraphics[width=0.95\linewidth]{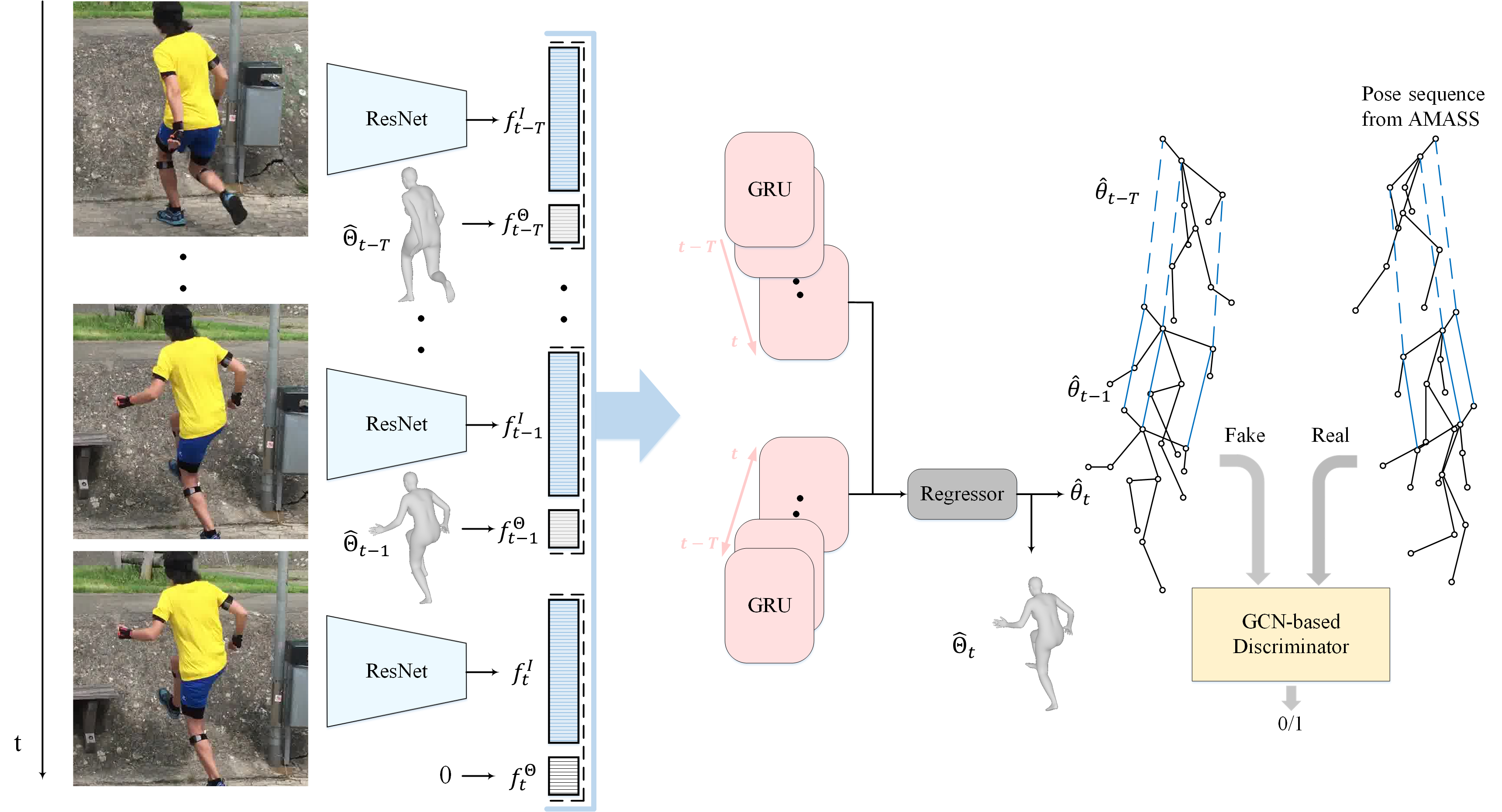}
  \caption{The overall architecture of our temporally embedded 3D human body pose and shape estimation (TePose). The network uses current frame, $T$ frames of previous frames, and previously predicted SMPL parameters as input to predict the human body pose and shape in the current frame. The temporal sequence of them is fed into two gated recurrent units (GRU) in time order. Graph convolution network (GCN) based motion discriminator is incorporated to include adversarial loss.}
  \label{fig:frame}
\end{figure*}
}

\newcommand{\gcn}{
\begin{figure}[t]
  \centering
  \includegraphics[width=1.0\linewidth]{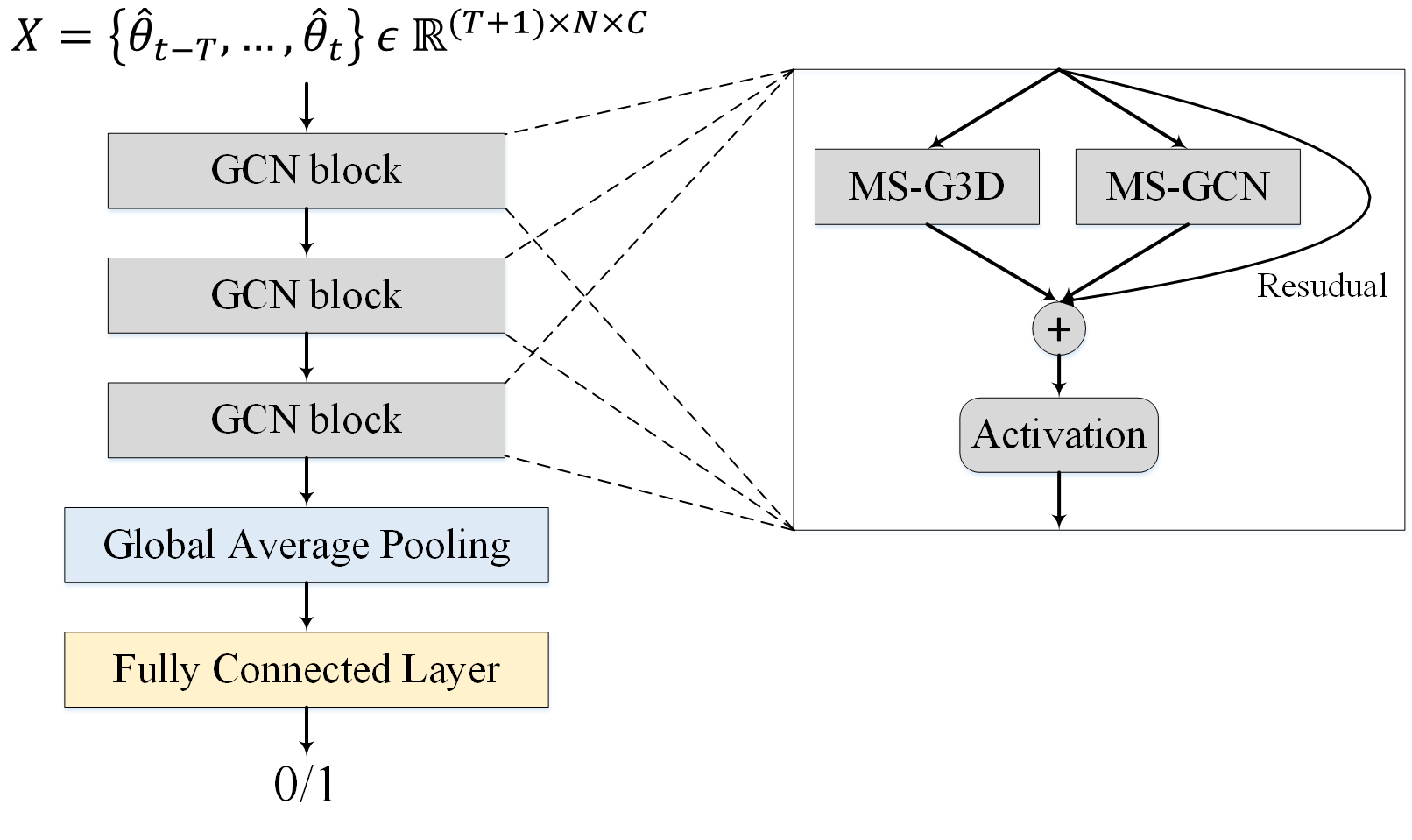}
  \caption{Motion discriminator architecture. It comprises three graph convolution network (GCN) blocks followed with global average pooling and one fully connected layer. Each GCN block adds the outputs of MS-G3D and MS-GCN units with residual connection. The number of output channels $C^{\prime}$ for the three GCN blocks are 64, 128 and 256 respectively.}
  \label{fig:gcn}
\end{figure}
}

\newcommand{\stages}{
\begin{figure}
  \centering
  \includegraphics[width=1\linewidth]{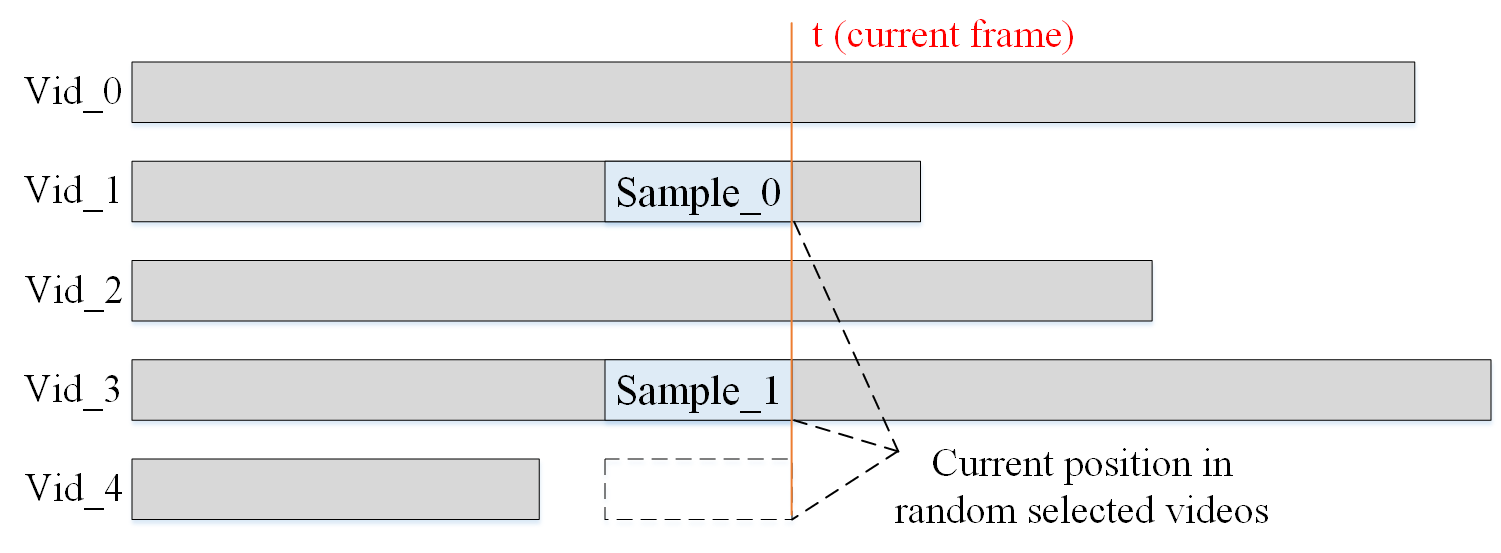}
  \caption{The sequential data loading strategy.}
  \label{fig:stages}
\end{figure}
}

\newcommand{\qual}{
\begin{figure*}[t]
  \centering
  \includegraphics[width=0.99\linewidth]{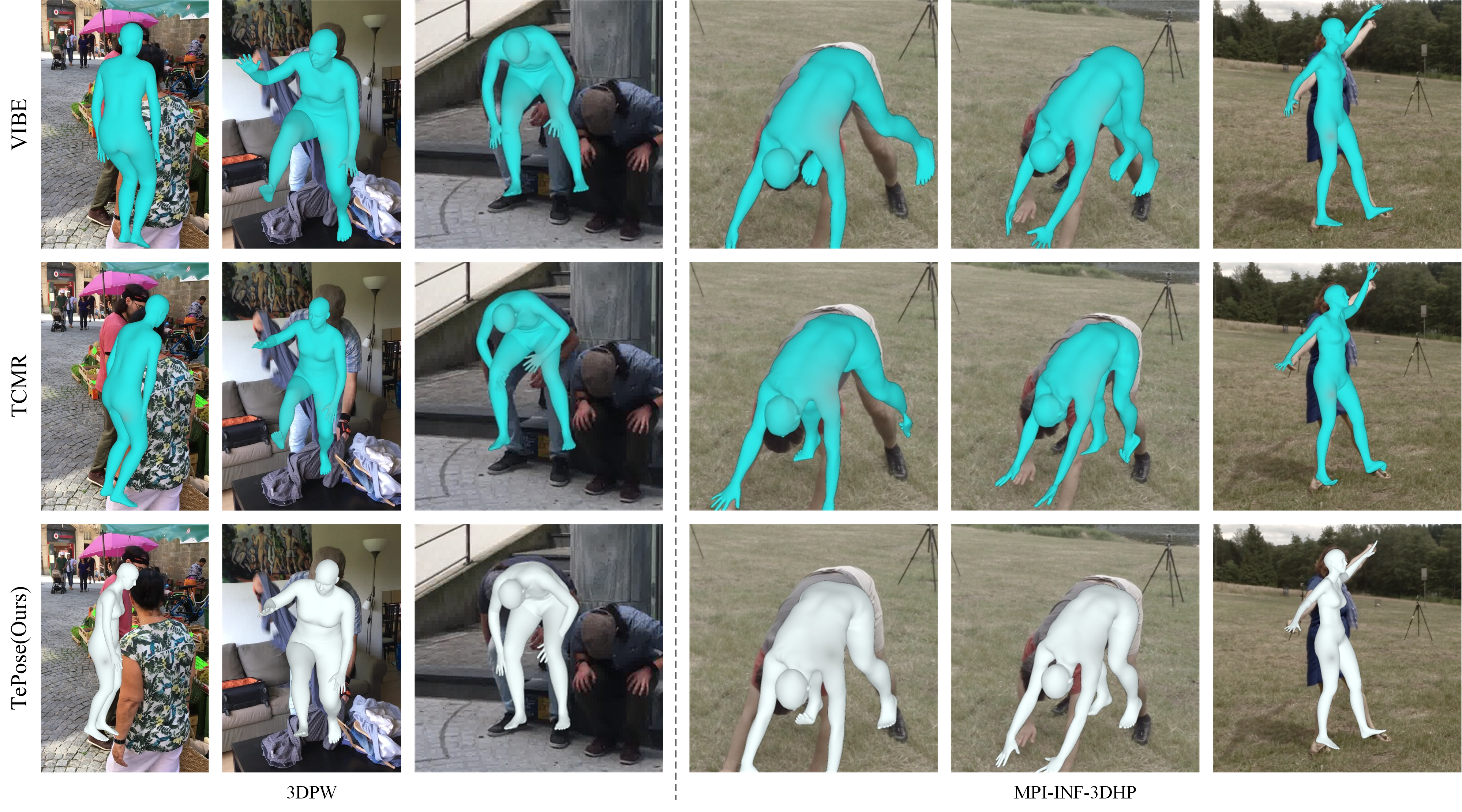}
  \caption{Qualitative comparisons between VIBE \cite{kocabas2020vibe} (top), TCMR \cite{choi2021beyond} (middle) and TePose (bottom) on in-the-wild sequences. The three frames on the left are from test set of 3DPW \cite{von2018recovering}, and the three frames on the right are from test set of MPI-INF-3DHP \cite{mehta2017monocular}.}
  \label{fig:qual}
\end{figure*}
}

\newcommand{\tabrescomparison}{
\begin{table*}[t]
\centering
{\begin{footnotesize}
\setlength\tabcolsep{3pt}
\begin{tabular}{c|c c c c|c c c|c c c|c}
\toprule
 & \multicolumn{4}{c}{3DPW} 
 & \multicolumn{3}{c}{MPI-INF-3DHP}
 & \multicolumn{3}{c}{Human3.6M}
\\
\cmidrule(r){2-11}
Methods
& \text{PA-MPJPE $\downarrow$}
& \text{MPJPE $\downarrow$}
& \text{MPVPE $\downarrow$}
& \text{Accel $\downarrow$}
& \text{PA-MPJPE $\downarrow$}
& \text{MPJPE $\downarrow$}
& \text{Accel $\downarrow$}
& \text{PA-MPJPE $\downarrow$}
& \text{MPJPE $\downarrow$}
& \text{Accel $\downarrow$}
& $T+1$
\\ 
\midrule
VIBE \cite{kocabas2020vibe}
& 57.6 & 91.9 & - & 25.4 & 68.9 & 103.9 & 27.3 & 53.3 & 78.0 & 27.3 & 16
\\
\cellcolor{Gray}MEVA \cite{luo20203d}$^*$
& \cellcolor{Gray}54.7 
& \cellcolor{Gray}86.9 
& \cellcolor{Gray}- 
& \cellcolor{Gray}11.6 
& \cellcolor{Gray}65.4 
& \cellcolor{Gray}96.4 
& \cellcolor{Gray}11.1 
& \cellcolor{Gray}53.2 
& \cellcolor{Gray}76.0 
& \cellcolor{Gray}15.3
& \cellcolor{Gray}90
\\
TCMR \cite{choi2021beyond}$^*$
& 52.7 & 86.5 & 102.9 & 7.1 & 63.5 & 97.3 & 8.5 & 52.0 & 73.6 & 3.9 & 16
\\
\textbf{TePose (Ours)}
& \cellcolor{Gray}\textbf{52.3} 
& \cellcolor{Gray}\textbf{84.6} 
& \cellcolor{Gray}\textbf{100.3} 
& \cellcolor{Gray}11.4 
& \cellcolor{Gray}\textbf{63.1}
& \cellcolor{Gray}\textbf{96.2}
& \cellcolor{Gray}16.7
& \cellcolor{Gray}\textbf{47.1}
& \cellcolor{Gray}\textbf{68.6}
& \cellcolor{Gray}12.1
& \cellcolor{Gray}6
\\
\bottomrule
\end{tabular}
\end{footnotesize}
}
\caption{Evaluation of state-of-the-art methods on 3DPW, MPI-INF-3DHP, and Human3.6M test datasets. All methods use 3DPW train set for training. $``*"$ denotes the methods cannot predict real-time. $``-"$ shows unavailable results. $T+1$ in the last column represents the number of input frames for each method.} 
\label{tab:tabrescomparison}
\end{table*}
}

\newcommand{\tabrescomparisonNoPW}{
\begin{table*}[t]
\centering
{\begin{footnotesize}
\setlength\tabcolsep{3pt}
\begin{tabular}{c c|c c c c|c c c|c c c}
\toprule
 & & \multicolumn{4}{c}{3DPW} 
 & \multicolumn{3}{c}{MPI-INF-3DHP}
 & \multicolumn{3}{c}{Human3.6M}
\\
\cmidrule(r){3-12}
 & Methods
& \text{PA-MPJPE $\downarrow$}
& \text{MPJPE $\downarrow$}
& \text{MPVPE $\downarrow$}
& \text{Accel $\downarrow$}
& \text{PA-MPJPE $\downarrow$}
& \text{MPJPE $\downarrow$}
& \text{Accel $\downarrow$}
& \text{PA-MPJPE $\downarrow$}
& \text{MPJPE $\downarrow$}
& \text{Accel $\downarrow$}
\\ 
\midrule
\multirow{5}{*}{\rotatebox[origin=c]{90}{Frame-based}}
& HMR \cite{kanazawa2018end}
& 76.7 & 130.0 & - & 37.4 & 89.9 & 124.2 & - & 56.8 & 88.0 & -
\\
& \cellcolor{Gray}{GraphCMR \cite{omran2018neural}}
& \cellcolor{Gray}70.2 
& \cellcolor{Gray}- 
& \cellcolor{Gray}- 
& \cellcolor{Gray}- 
& \cellcolor{Gray}- 
& \cellcolor{Gray}- 
& \cellcolor{Gray}- 
& \cellcolor{Gray}50.1 
& \cellcolor{Gray}- 
& \cellcolor{Gray}-
\\
& SPIN \cite{kolotouros2019learning}
& 59.2 & 96.9 & 116.4 & 29.8 & 67.5 & 105.2 & - & 41.1 & - & 18.3
\\
& \cellcolor{Gray}I2L-MeshNet \cite{moon2020i2l}
& \cellcolor{Gray}57.7 
& \cellcolor{Gray}93.2 
& \cellcolor{Gray}110.1 
& \cellcolor{Gray}30.9 
& \cellcolor{Gray}- 
& \cellcolor{Gray}- 
& \cellcolor{Gray}- 
& \cellcolor{Gray}41.1 
& \cellcolor{Gray}\textbf{55.7} 
& \cellcolor{Gray}13.4
\\
& Pose2Mesh \cite{choi2020pose2mesh}
& 58.3 & \textbf{88.9} & \textbf{106.3} & 22.6 & - & - & - & 46.3 & 64.9 & 23.9
\\
\midrule
\multirow{6}{*}{\rotatebox[origin=c]{90}{Temporal}}
& \cellcolor{Gray}HMMR \cite{kanazawa2019learning}
& \cellcolor{Gray}72.6 
& \cellcolor{Gray}116.5 
& \cellcolor{Gray}139.3 
& \cellcolor{Gray}15.2 
& \cellcolor{Gray}- 
& \cellcolor{Gray}- 
& \cellcolor{Gray}- 
& \cellcolor{Gray}56.9 
& \cellcolor{Gray}- 
& \cellcolor{Gray}-
\\
& Doersch \textit{et al.} \cite{doersch2019sim2real}
& 74.7 & - & - & - & - & - & - & - & - & -
\\
& \cellcolor{Gray}Sun \textit{et al.} \cite{sun2019human}
& \cellcolor{Gray}69.5 
& \cellcolor{Gray}- 
& \cellcolor{Gray}- 
& \cellcolor{Gray}- 
& \cellcolor{Gray}- 
& \cellcolor{Gray}- 
& \cellcolor{Gray}- 
& \cellcolor{Gray}42.4 
& \cellcolor{Gray}\textbf{59.1} 
& \cellcolor{Gray}-
\\
& VIBE \cite{kocabas2020vibe}
& 56.5 & 93.5 & 113.4 & 27.1 & 63.4 & \textbf{97.7} & 29.0 & 41.5 & 65.9 & 18.3
\\
& \cellcolor{Gray}\textbf{TePose (Ours)}
& \cellcolor{Gray}\textbf{56.1} 
& \cellcolor{Gray}93.9 
& \cellcolor{Gray}115.9 
& \cellcolor{Gray}\textbf{11.7}
& \cellcolor{Gray}\textbf{62.9} 
& \cellcolor{Gray}99.5 
& \cellcolor{Gray}\textbf{17.2} 
& \cellcolor{Gray}\textbf{41.2} 
& \cellcolor{Gray}61.6 
& \cellcolor{Gray}\textbf{12.0}
\\
\bottomrule
\end{tabular}
\end{footnotesize}
}
\caption{Evaluation of state-of-the-art methods on 3DPW, MPI-INF-3DHP, and Human3.6M test datasets. All methods do not use
3DPW \cite{von2018recovering} train set for training. $``-"$ shows unavailable results.} 
\label{tab:tabrescomparisonNoPW}
\end{table*}
}

\newcommand{\tabablation}{
\begin{table}[h]
\centering
\small
{
\setlength\tabcolsep{2pt}
\begin{tabular}{c c c c | c c}
\toprule
2-GRU & Embedded & Seq & Adv
& \text{PA-MPJPE $\downarrow$}
& \text{MPJPE $\downarrow$}
\\
\midrule
\ding{55} & \ding{55} & \ding{55} & \ding{55} 
& 42.6 & 64.4
\\
\ding{51} & \ding{55} & \ding{55} & \ding{55} 
& 41.7 & 62.9
\\
\ding{51} & \ding{51} & \ding{51} & \ding{55} 
& 41.5 & 62.2
\\
\ding{51} & \ding{51} & \ding{51} & \ding{51} 
& \textbf{41.2} & \textbf{61.6}
\\
\bottomrule
\end{tabular}
}
\caption{Ablation experiments with different combinations of proposed modules. Models are trained on MPI-INF-3DHP, Human3.6M and InstaVariety while all of them are evaluated on the test set of Human3.6M. 'Embedded' represents the temporally embedded encoder. 'Seq' denotes the sequential data loading strategy. 'Adv' denotes the adversarial learning module using GCN-based motion discriminator.} 
\label{tab:tabablation}
\vspace{-.2in}
\end{table}
}

\definecolor{Gray}{gray}{0.9}

\newcommand{\figref}[1]{Figure~\ref{fig:#1}}
\newcommand{\tabref}[1]{Table~\ref{tab:#1}}

\def\wacvPaperID{1083} 

\wacvalgorithmstrack   

\wacvfinalcopy 

\ifwacvfinal
\usepackage[breaklinks=true,bookmarks=false]{hyperref}
\else
\usepackage[pagebackref=true,breaklinks=true,colorlinks,bookmarks=false]{hyperref}
\fi

\begin{document}

\title{Live Stream Temporally Embedded 3D Human Body Pose and Shape Estimation}

\author{Zhouping Wang and Sarah Ostadabbas\\
Augmented Cognition Lab (ACLab), Northeastern University, Boston, MA, USA\\
{\tt\small \{zpwang,ostadabbas\}@ece.neu.edu}
}

\maketitle
\thispagestyle{empty}

\begin{abstract}
  3D Human body pose and shape estimation within a temporal sequence can be quite critical for understanding human behavior. Despite the significant progress in  human pose estimation  in the recent years, which are often based on single images or videos, human motion estimation on live stream videos is still a rarely-touched area considering its special requirements for real-time output and temporal consistency. To address this problem, we present a temporally embedded 3D human body pose and shape estimation (TePose) method to improve the accuracy and temporal consistency of pose estimation in live stream videos. TePose uses previous predictions as a bridge to feedback the error for better estimation in the current frame and to learn the correspondence between data frames and predictions in the history. A multi-scale spatio-temporal graph convolutional network is presented as the motion discriminator for adversarial training using datasets without any 3D labeling. We propose a sequential data loading strategy to meet the special start-to-end data processing requirement of live stream. We demonstrate the importance of each proposed module with extensive experiments. The results show the effectiveness of TePose on widely-used human pose benchmarks  with state-of-the-art performance. \footnote{Code available at \href{https://github.com/ostadabbas/TePose}{https://github.com/ostadabbas/TePose}.}.
  
\end{abstract}

\section{Introduction}
Estimating human pose and shape is a crucial yet challenging task for many human-centric 3D computer vision applications \cite{boulic1997integration,presti20163d,luvizon2017learning,li2020detailed,li2020pastanet}. With the recent developments in this area, model-based methods are designed to estimate 3D human pose and shape with a few parameters. Despite the improvements of accuracy and robustness for single images and video clips, live stream, as a major realtime processing media, has lagged to be explored for improving the accuracy and temporal consistency of human pose and shape estimation.

Single image-based methods \cite{bogo2016keep,tan2017indirect,lassner2017unite,pavlakos2019expressive,pavlakos2018learning,omran2018neural,kanazawa2018end,guler2019holopose,georgakis2020hierarchical,li2021hybrik} have widely been explored to recover 3D human pose and shape. Some of them have achieved impressive performance on the standard of spatial errors. However, the errors of estimation for each image are highly independent, even being applied to consecutive frames of a video or live stream. Minor spatial fluctuations for estimations in a frame sequence will be conspicuous for human observation as the fluctuations can be in different directions. The lack of temporal information makes single image-based methods much harder to get temporally consistent with stable pose and shape estimation.

Some recent works have been proposed to extend the single image-based methods to the video cases \cite{kanazawa2019learning,kocabas2020vibe,luo20203d}. A sequence of frames is fed into pretrained single image-based 3D human pose and shape estimation networks to get the static features for each frame \cite{kanazawa2018end,kolotouros2019learning}. Then a temporal encoder is utilized to extract temporal features from these static features. Finally, a mesh parameter regressor outputs the parameters of a template-based model, called skinned multi-person linear (SMPL)  \cite{loper2015smpl,pavlakos2019expressive}, for each frame by inputting temporal features. Choi \textit{et al.} \cite{choi2021beyond} improved the temporal consistency of 3D human motion by leveraging the temporal information from past and future frames excluding the static feature of the current frame. However, using information of future frames for prediction is not allowed in live stream scenarios. The temporal information of these video-based methods is limited by the length of the input sequence. To extract affluent temporal relevance to assist stable pose and shape recovery, the input video is usually quite long, which impedes their real-time applications.

We propose that the pose and shape predictions for previous frames can be a bridge to bring history information into account for current frame prediction, called temporally embedded 3D human body pose and shape estimation (TePose). TePose uses both the video frames and their corresponding pose and shape from previous predictions as the input for the current frame prediction. All the static features of input frames except current frame are concatenated with their pose and shape from previous predictions. A temporal encoder and subsequent regressor will predict the pose and shape for current frame using these input information. Given that the static features are extracted by network trained on large datasets, the SMPL parameters inherited in the static features will have sample-based shift. The previous pose and shape predictions will play a role as feedback for the temporal encoder to learn sample-based correspondence between previous frames with the pose and shape predictions for fixing this shift.

Compared to methods using only video frames as model input, our model needs to use the human body pose and shape predictions from previous frames as input for prediction at the current frame. And the predicted pose and shape as the input are inaccurate especially at the beginning of the network training, which could make the convergences of our network even harder. In order to address this problem, we present a sequential data loading strategy and use mixed sources of pose and shape as the input for prediction. The proportion of input using ground truth pose and shape helps the model to converge at the beginning of training. The pose and shape predictions for previous frames come back as part of the input to make the network gradually fit into the real-world scenarios.

Kocabas \textit{et al.}  introduced in-the-wild datasets for the training of 3D human pose and shape estimation network by using adversarial training strategy \cite{kocabas2020vibe}. The adversarial loss can make up for 3D prediction evaluation whenever there is no 3D labeling for in-the-wild data. Graph neural network has long been proved to have strong ability for classification applications on different modalities of graphs including human pose skeleton. In this work, we use multi-scale spatial-temporal graph convolutional network as the motion discriminator to distinguish real human skeleton sequences against predicted ones. The real skeletons from AMASS dataset \cite{mahmood2019amass} are used for training our motion discriminator.

In summary, this work introduces temporally embedded 3D human body pose and shape estimation (TePose), with the key contributions as following: 

\begin{itemize}
    \item presenting a temporally embedded human body pose and shape recovery framework using previous predictions as a bridge to feedback the error for current frame and to learn the correspondence between data frames and predictions in history. 
    \item developing a sequential data loading strategy with mixed mesh parameter inputs to help the proposed model fitting from ideal cases to real-world cases. 
    \item introducing a graph convolutional network (GCN) based motion discriminator to provide 3D prediction evaluation in absence of 3D labeling.
    \item comparing different temporal embedded architectures for 3D human body pose and shape recovery. 
    \item and achieving the state-of-the-art performance on widely-used benchmarks for live stream human body pose and shape recovery. 
\end{itemize}

\section{Related Work}
Previous methods predict the human body pose and shape from a small set of video or live stream frames, which is the current frame for single image-based methods and several adjacent frames for video clip-based methods. The independence of these inputs make it hard to further improve the temporal consistency and accuracy for the prediction. Using more frames from longer period of time as input is beneficial for temporal information learning, however it is inefficient and time-consuming. Below, we give an overview of the state-of-the-art works in each category.

\paragraph{3D human body pose and shape from a single image.}
Model-based methods that predict the parameters of predefined human body model \cite{anguelov2005scape,loper2015smpl,pavlakos2019expressive,yang2021s3} have been widely used for 3D human body pose and shape estimation. The mapping from 2D monocular image to 3D parameters is an ill-posed problem and lacks of in-the-wild 3D supervision. Bogo \textit{et al.} \cite{bogo2016keep} proposed SMPLify by optimizing the SMPL parameters to fit the estimated 2D keypoints, which relieves the need for manual assist \cite{guan2009estimating}. This kind of self-supervision strategy has further been extended to fit silhouette \cite{tan2017indirect,lassner2017unite} and 2D features for face, hands, and feet \cite{pavlakos2019expressive}. Kolotouros \textit{et al.} introduced a CNN-based regressor into SMPLify framework to form a self-improving system \cite{kolotouros2019learning}. Some works presented intermediate representations including keypoint heatmaps, silhouette \cite{pavlakos2018learning}, and semantic segmentation \cite{omran2018neural} to split the whole mapping into two separate parts, which can be trained with broader supervisions. Kanazawa \textit{et al.} \cite{kanazawa2018end} used adversarial loss to supervise the human mesh recovery with an end-to-end framework comprised of convolutional encoder and SMPL parameter regressor.
HoloPose \cite{guler2019holopose} presented body-part-based prior and iterative fitting for DensePose \cite{guler2018densepose} and 2D/3D joints to improve the regression. Georgakis \textit{et al.} \cite{georgakis2020hierarchical} provided hierarchical kinematic prior to improve the repression performance. Li \textit{et al.} proposed the HybrIK \cite{li2021hybrik} by using the hybrid inverse kinematics solution to bridge the gap between body mesh recovery and 3D keypoint estimation so as to get regression improvement.

There are also some model-free methods directly recovering the human body mesh instead of regressing body model parameters. Kolotouros \textit{et al.} \cite{kolotouros2019convolutional} directly regressed the vertex coordinates for a mesh template using graph convolutional network (GCN) \cite{kipf2016semi}. Saito \textit{et al.} \cite{saito2019pifu} aligned pixels of 2D images with the global context of their corresponding 3D object to infer both 3D surface and texture from a single image. Moon and Lee \cite{moon2020i2l} predicted the per-lixel (line+pixel) likelihood on 1D heatmaps for each mesh vertex coordinate instead of directly regressing the parameters. Choi \textit{et al.} \cite{choi2020pose2mesh} estimated the 3D coordinates of human mesh vertices directly from the 2D human pose using a GCN-based system. \textbf{Shortcomings--}The single-image based methods have achieved impressive per-frame accuracy. However, as they do not utilize temporal information, lacking of temporal consistency restricts their performance in real-scenario video or live stream applications.

\paragraph{3D human body pose and shape from video.}
Some recent approaches have extended the SMPLify to exhibit a spatial-temporal manner. Tung \textit{et al.} \cite{tung2017self} used self-supervised losses driven by differentiable keypoint, segmentation, and motion reprojection errors, against detected 2D keypoints, 2D segmentation and 2D optical flow, respectively. Kanazawa \textit{et al.} \cite{kanazawa2019learning} used ResNet-based encoder \cite{he2016identity} to learn human motion dynamics by supervising the dynamics prediction over time. Arnab \textit{et al.} \cite{arnab2019exploiting} exploited the temporal context with bundle adjustment using the internet videos annotated by modified SMPLify. Doersch and Zisserman \cite{doersch2019sim2real} proved that synthetic data with cues about the person's motion, notably
as optical flow and the motion of 2D keypoints, can help improving the performance for human mesh recovery. Sun \textit{et al.} \cite{sun2019human} proposed a skeleton-disentangling based framework to reduce the complexity and decouple the skeleton, a self-attention based temporal convolution network to efficiently exploit the short and long-term temporal cues, and a shuffled frame order recovery strategy to promote the learning of motion dynamics. Luo \textit{et al.} \cite{luo20203d} proposed a motion estimation method via variational autoencoder (MEVA) using auto-encoder-based motion compression and a residual representation learned through motion refinement. Kocabs \textit{et al.} presented the VIBE \cite{kocabas2020vibe} approach by introducing an adversarial learning framework with self-attention mechanism that leverages AMASS dataset \cite{mahmood2019amass} to supervise the temporal human body pose and shape regression networks. Choi \textit{et al.} proposed a temporally consistent mesh recovery (TCMR) \cite{choi2021beyond} system to recover the mesh for the middle frame of a video clip by fusing different kinds of temporal encoders and removing the temporal-inconsistent effect getting from residual connection. Lee \textit{et al.} \cite{lee2021uncertainty} proposed to employ a view-invariant probabilistic encoder that can present 2D pose features as distribution of uncertainty in 2D space and a decoder that divides the body into five different local regions to estimate the 3D motion dynamics of each region.  \textbf{Shortcomings--}However, both TCMR and this method need future frames as input for current frame estimation.  In addition, most of video-based systems need long sequence of frames as input to get enough temporal information.

\paragraph{Graph neural network (GNN) for human pose skeleton.}
In the last decade, GNNs have been developed with many different forms to extract features from data in a graph \cite{bruna2013spectral,hamilton2017inductive,kipf2016semi,velivckovic2017graph,xu2018powerful,gao2019graph}. GNNs can be roughly divided into spectral GNNs \cite{bruna2013spectral,kipf2016semi} or spatial GNNs \cite{hamilton2017inductive}. Graph convolution network (GCN) \cite{kipf2016semi} was initially proposed as a first-order approximation for localized spectral convolutions. Its simplicity of structure and calculation makes it popular for graph-involved applications \cite{wang2018videos,yu2017spatio}.
Multi-scale spatial GNN \cite{liao2019lanczosnet} has been proposed to capture non-local features. Spatial-temporal GCN \cite{yan2018spatial} uses human skeleton sequences as input for action classification task, which has been improved by Liu \textit{et al.} \cite{liu2020disentangling} after extending the Multi-scale spatial-temporal graph convolution network (MS-GCN) \cite{chaolong2018spatio} into its disentangled version known as MS-G3D.

\section{Introducing TePose}

The overall architecture of temporally embedded 3D human body pose and shape estimation (TePose) is shown in \figref{frame}. It mainly comprises two parts, one a temporally embedded encoder and the other one a series of GCN-based motion discriminators. When a live stream or video sequence is given to estimate the human body pose and shape in each of its frame, TePose realizes this goal following a frame-by-frame manner, as described below.

\textbf{Problem Formulation--}We define the just-received frame for live stream or the frame in estimation for video as current frame $I_t$. The $T$ frames before current frames, $\{I_{t-T},...,I_{t-1}\}$, and their corresponding predicted SMPL parameters, $\{\Theta_{t-T},...,\Theta_{t-1}\}$, are used to help  the human body pose and shape estimation for the current frame. At the start of each live stream or video, we utilize the VIBE \cite{kocabas2020vibe}, which is a video-based inference model for body pose and shape estimation, to provide the SMPL parameters for the first $T$ frames. Now, we introduce the detailed description of each component.

\framework

\subsection{Temporally Embedded Encoder}
Given a sequence of RGB frames $\{I_{t-T},...,I_{t-1},I_t\}$ and previous SMPL parameter predictions $\{\Theta_{t-T},...,$ $\Theta_{t-1}\}$, a pretrained ResNet is used to extract the static image features $\{f_{t-T}^I,...,f_{t}^I\}$ with dimension $\mathbb{R}^{2048}$ for each frame \cite{kolotouros2019learning}. The previously predicted SMPL parameters $\{\Theta_{\bullet}\}$ for each frame, which include camera parameters $\pi_{\bullet}\in\mathbb{R}^3$, pose parameters $\theta_{\bullet}\in\mathbb{R}^{72}$ and shape parameters $\beta_{\bullet}\in\mathbb{R}^{10}$, are recognized as another input to concatenate with the image features as:
\begin{equation}
   f_i =
   \text{Cat}(f_i^I,f_i^{\hat{\Theta}}),i=t-T,...,t-1.
   \label{eq:fea}
\end{equation}

The static image features of current frame are concatenated with zeros vector to match the dimension with other frames. Then the concatenated features $\{f_{t-T},...,f_t\}$ are sent into two gated recurrent units (GRU) \cite{cho2014learning} in the time order to compute hidden states $\{g_{t-T},...,g_t\}$. One of the GRU is uni-directional and another one is bi-directional. The initial hidden states of these GRUs are set to zeros. The output hidden states of GRUs for current frame $g_t$ are inputted into a SMPL parameter regressor \cite{kolotouros2019learning}. Following previous work \cite{kocabas2020vibe,choi2021beyond}, The regressor is initialized with the mean pose and shape. Finally, the regressor outputs the predictions for the SMPL parameters $\hat{\Theta}_i=\{\hat{\pi}_i,\hat{\theta}_i,\hat{\beta}_i\}$, 2D joints, 3D joints and the SMPL mesh coordinates $\mathcal{M}(\hat{\theta}_i,\hat{\beta}_i)\in\mathbb{R}^{6890\times3}$. A human body pose and shape model with neutral gender is used as in previous works. During the training process, we get two different predictions for features extracted from the two GRUs. Both of the predictions are used for calculating loss. However, we average the output features of two GRUs during evaluation so that we can get identical prediction for each frame.

\subsection{GCN-based Motion Discriminator}
One of the major problems faced by human body pose and shape estimation is the lack of in-the-wild datasets with accurate 3D annotations. Indoor datasets with 3D annotations have obvious differences compared with in-the-wild ones for data distributions including backgrounds and variety of subjects. The pseudo labels for some in-the-wild datasets are not reliable. To relieve this problem, we introduce adversarial loss to constrain the 3D human pose estimation. A multi-scale spatial-temporal graph convolutional network is used to distinguish the predicted 3D human skeleton sequences with the ones from human motion capture datasets. Our temporally embedded encoder is trained to generate 3D skeletons as real as possible to cheat the motion discriminator in an adversarial way.

The network structure of GCN-based motion discriminator is visualized in \figref{gcn}. One of the outputs of the temporally embedded encoder introduced in last section is the predicted human skeleton sequence $\{\hat{\theta}_{t-T},...,\hat{\theta}_{t}\}$. We unify the predicted skeleton sequence with real ones from motion capture datasets as $X\in\mathbb{R}^{(T+1)\times N\times C}$, where $N$ is the number of joints and $C$ is the coordinate dimension for each joint. The skeleton sequences are firstly processed by three consecutive GCN blocks. Then the global average pooling layer and fully connected layer are used to output the final decision if the input skeleton sequence is real or generated. For each GCN block, graph features are processed by multi-scale graph convolutional network (MS-GCN) \cite{chaolong2018spatio} and multi-scale graph 3D convolutional network (MS-G3D) \cite{liu2020disentangling}. Then their outputs are added with residual connection. Finally, the graph features are processed with activation function to get the output for next step.

\gcn

The MS-GCN is used to apply graph convolution on the human skeleton for each frame, namely spatial dimension. It can be represented as: 
\begin{equation}
   X_t^\prime =
   \sigma\left[\sum_{k=0}^K\tilde{D}_{(k)}^{-\frac{1}{2}}\tilde{A}_{(k)}\tilde{D}_{(k)}^{-\frac{1}{2}}X_{t}W_{(k)}\right].
   \label{eq:gcn}
\end{equation}
The $\sigma$ is the activation function, and the $\tilde{A}_{(k)}$ is the $k$-adjacency matrix for the graph of human skeleton,
\begin{equation}
   \left[\tilde{A}_{(k)}\right]_{m,n} =
   \begin{cases}
     1 & \text{if \ $d(X_t^{(m)},X_t^{(n)})=k,$}\\
     1 & \text{if \ $m=n,$}\\
     0 & \text{otherwise,}
   \end{cases}
\end{equation}
where the $d(X_t^{(m)},X_t^{(n)})$ represents the least number of hops to connect nodes $X_t^{(m)}$ and $X_t^{(n)}$, and the $k$ denotes the kernel scale on the spatial graph. $\tilde{D}_{(k)}$ is the diagonal degree matrix of $\tilde{A}_{(k)}$ to get normalized $k$-adjacency matrix $\tilde{D}_{(k)}^{-\frac{1}{2}}\tilde{A}_{(k)}\tilde{D}_{(k)}^{-\frac{1}{2}}$ \cite{kipf2016semi}. And the $W_{(k)}\in\mathbb{R}^{C\times C^{\prime}}$ denotes the learnable weight matrix for the graph convolution.

The MS-G3D is used to apply graph convolution on the human skeletons in a time range $\tau$, namely both on spatial and temporal dimensions. It can be represented as:
\begin{footnotesize}
\begin{equation}
   X_{t\pm\tau/2}^\prime =
   \sigma\bigg\{\sum_{k=0}^K\left[\tilde{D}_{(k)}^{(\tau)}\right]^{-\frac{1}{2}}\tilde{A}_{(k)}^{(\tau)}\left[\tilde{D}_{(k)}^{(\tau)}\right]^{-\frac{1}{2}}X_{t\pm\tau/2}W_{(k)}\bigg\}.
   \label{eq:g3d}
\end{equation}
\end{footnotesize}
Compared with Eq. \ref{eq:gcn}, the input skeleton data in Eq. \ref{eq:g3d} is extended to be spatial-temporal $X_{t\pm\tau/2}\in\mathbb{R}^{\tau\times N\times C}$. The adjacency matrix for MS-G3D is also extend to be: 
\begin{equation}
   \tilde{A}_{(k)}^{(\tau)} =
   \begin{bmatrix}
   \tilde{A}_{(k)} & \dots & \tilde{A}_{(k)}\\
   \vdots & \ddots & \vdots\\
   \tilde{A}_{(k)} & \dots & \tilde{A}_{(k)}
   \end{bmatrix}_{\tau\times \tau}.
\end{equation}

$\tilde{D}_{(k)}^{(\tau)}$ is the diagonal degree matrix for $\tilde{A}_{(k)}^{(\tau)}$. By using zero padding at temporal dimension, the output skeleton data of each MS-G3D share the same spatial and temporal dimensions with the input $X^{\prime}\in\mathbb{R}^{(T+1)\times N\times C^{\prime}}$.

\subsection{Loss Functions}
The network is trained on both the datasets with 3D annotations and the ones without 3D annotations. Motivated by previous works \cite{kocabas2020vibe}, the total loss of proposed model is:
\begin{equation}
   L =
   L_{2D}+\mathbb{1}_{3D}L_{3D}+\mathbb{1}_{\Theta}L_{\Theta}+(1-\mathbb{1}_{\Theta})L_{adv},
\end{equation}
where $\{\mathbb{1}_{3D},\mathbb{1}_{\Theta}\}$ are indicators to show if 3D keypoint labels and SMPL parameter labels exist respectively. A sample with 3D keypoint label will have loss function with $\mathbb{1}_{3D}=1$, otherwise 0. And a sample with SMPL parameter label will have loss function with $\mathbb{1}_{\Theta}=1$, otherwise 0. We use the $L2$ losses for 2D, 3D and SMPL losses:
\begin{equation}
   L_{3D} =
   \|\hat{X}_t-X_t\|_2, \nonumber
\end{equation}
\begin{equation}
   L_{2D} =
   \|\hat{x}_t-x_t\|_2, \nonumber
\end{equation}
\begin{equation}
   L_{\Theta} =
   \|\hat{\theta}_t-\theta_t\|_2+\|\hat{\beta}_t-\beta_t\|_2,
\end{equation}
where $x_t$ is the ground truth for 2D keypoints, and $\hat{x}_t=s\Pi RX_t+\iota$ is the 2D keypoint predictions deriving from 3D reprojection. $\left[s,\iota,R\right]$ are scale, transition and global rotation matrix from camera parameters. $\Pi$ represents the orthographic projection. The adversarial loss can be represented as
\begin{equation}
   L_{adv} =
   \left[\mathcal{D}_M\left(\hat{X}\right)-1\right]^2,
\end{equation}
where $\mathcal{D}_M$ denotes the proposed GCN-based motion discriminator. And the cross-entropy loss is used for training this motion discriminator: 
\begin{small}
\begin{equation}
   L_{\mathcal{D}_M} =
   \mathbb{E}_X\left[\left(\mathcal{D}_M(X)-1\right)^2\right]+\mathbb{E}_{\hat{X}}\left[\left(\mathcal{D}_M(\hat{X})\right)^2\right].
\end{equation}
\end{small}

\subsection{Sequential Data Loading Strategy}
In our approach, we have to use both the previous image frames and their corresponding SMPL parameter predictions as inputs to train the network for the pose and shape estimation of current frame. The sequential training moving from the start to the end of each video is needed in this scenario as only the predictions for previous frames have got then the subsequent frames can be processed. Different from the common data loading method for machine learning, video sequences in our training procedure are no longer selected in a totally random way. We use the current training iteration $j$ to get the current frame position $t=mod(j,H)$, where $H$ is defined to be the maximum video length we select to reach. The sequential data loading strategy of proposed method is shown in \figref{stages}. The long videos are kept selecting with a random way, and the samples are in the current frame positions of each selected long video. If there are selected long video shorter than the $H$, we do not include these ones into the batch for training and squeeze the batch size. In order to better control the convergence of our network, we use a mixed way to select the source of previous human body pose and shape. There is a probability of $\gamma$ to use previously predicted SMPL parameters as in Eq. \ref{eq:fea} and a probability of $1-\gamma$ to use ground truth labels for network input as
\begin{equation}
   f_i =
   \text{Cat}(f_i^I,f_i^{\Theta}),i=t-T,...,t-1.
   \label{eq:feaGT}
\end{equation}

\stages

\subsection{Implementation Details}
We follow some of the settings as the previous works \cite{kocabas2020vibe,choi2021beyond}. The frame rate of video sequences are 25--30 frames per second. The backbone and regressor in the model is initialized with pretrained SPIN model \cite{kolotouros2019learning}. All the input human frames are cropped using ground truth bounding boxes and resized to 224$\times$224 \cite{kanazawa2018end,kocabas2020vibe,kolotouros2019learning,kolotouros2019convolutional}. Occlusion augmentation is applied to the cropped frames for better generalization of the model \cite{sarandi2018robust}. 2-layer GRUs \cite{cho2014learning} with hidden size of 1024 are used in our temporally embedded encoder. The regressor has 2 fully-connected layers with 1024 neurons for each layer, and a final layer to output SMPL parameters\cite{kolotouros2019learning}. To reduce the time for waiting buffer frames and computational complexity for prediction, we choose a small temporal length $T+1=6$. The number of kernel scales for all the MS-GCN and MS-G3D units in the discriminator are set to be $K=13$ and $K=6$ respectively. And the temporal window size $\tau$ is set to be 3 for all the MS-G3D units. The probability of using previously predicted SMPL parameters as input is set to $\gamma=0.9$. The network is trained for 80 epochs with one NVIDIA Tesla V100 SXM2 GPU. During each epoch, 1/8 of videos from 3D human body pose estimation datasets are used for training. Maximum video length $H$ is set to $505$. The learning rates for predictor and discriminator are initialized to $5^{-5}$ and $1^{-4}$, and reduced by a factor of 10 when the 3D pose accuracy does not improve after every 8 epochs. Adam optimizer \cite{kingma2014adam} is used for updating the network weights with a mini-batch size of 32.

The number of learnable parameters for the 2-layer GRUs in our system is $O(18(len(g_t)^2+len(g_t)\cdot len(f_t)+len(g_t)))=O(18\times(2048^2+2048\times(2048+75)+2048))$. The number of learnable parameters for each MS-GCN is $O(K_1 C C^\prime)$, where $K_1=13$ is the number of kernel scales. $C$ and $C^\prime$ are the numbers of input and output channels. The number of learnable parameters for each MS-G3D is $O(K_2 C C^\prime)$ while $K_2=6$. The total number of learnable parameters for proposed discriminator is $O((K_1+K_2)(CC_1^\prime+C_1^{\prime}C_2^{\prime}+C_2^{\prime}C_3^{\prime})=O((13+6)(3\times 64+64\times128+128\times256)))$.

\tabrescomparison
\tabrescomparisonNoPW
\section{Experiments}
\label{others}

\subsection{Datasets and Evaluation Metrics}
We use both 3D and 2D human body pose estimation datasets for our model training. 3DPW \cite{von2018recovering} is an in-the-wild dataset with 3D mesh labels, and Human3.6M \cite{ionescu2013human3} also has SMPL parameter labels obtained from Mosh \cite{loper2014mosh}. MPI-INF-3DHP \cite{mehta2017monocular} is 3D dataset consisting of both indoor and complex outdoor scenes. InstaVariety \cite{kanazawa2019learning} is a dataset with pseudo 2D keypoints. PoseTrack \cite{andriluka2018posetrack} is the only ground-truth 2D video datasets we use. 3D datasets and 2D datasets are mixed together for supervision. The ratio between the numbers of 3D and 2D data samples is set to $4:6$. Pose sequences from AMASS \cite{mahmood2019amass} are used as real samples for adversarial training. The test sets of 3DPW, Human3.6M and MPI-INF-3DHP are also used for evaluation.

For performance evaluation, we use both per-frame-based metrics, including mean per joint position error (MPJPE), Procrustes-aligned MPJPE (PA-MPJPE) and mean per vertex position error (MPVPE), and temporal-based metrics, acceleration error (ACCEL) \cite{kanazawa2019learning}, to calculate the prediction errors. These position errors are measured in millimeter between the estimated and groundtruth 3D coordinates after aligning the root joint.

\subsection{Comparison with the State-of-the-art}
Based on the work of Luo et al. \cite{luo20203d}, we compare our proposed Tepose approach with state-of-the-art models in two different training settings. In \tabref{tabrescomparison}, all the models are trained with multiple datasets including 3DPW train set, while tested on the test sets of 3DPW, MPI-INF-3DHP and Human3.6M respectively. Specifically, for our model evaluating on 3DPW test set, we use train sets of 3DPW, MPI-INF-3DHP, Human3.6M and InstaVariety for training. only 3DPW with SMPL parameter labels is applied with SMPL loss $L_{\Theta}$. The samples from MPI-INF-3DHP and Human3.6M without true mesh labels use adversarial loss $L_{adv}$ instead.
For our model evaluating on MPI-INF-3DHP and Human3.6M test sets, it is trained with 3DPW, MPI-INF-3DHP, Human3.6M, InstaVariety and PoseTrack train sets. And the SMPL parameters of Human3.6M obtained from Mosh \cite{loper2014mosh} are added for supervision through SMPL loss $L_{\Theta}$. As shown in the table, proposed TePose has best performance on all the per-frame-based metrics. MEVA \cite{luo20203d} needs at least 90 frames as model input, and TCMR \cite{choi2021beyond} predicts the body pose and shape of middle frame from 16-frame input, which impede them from real-time applications considering they need future frames as input. In contrast, our approach can infer without using future frames, and only 6 frames are needed as model input. Comparing with VIBE \cite{kocabas2020vibe}, which can also do real-time prediction, proposed TePose is better on both per-frame-based metrics and temporal-based metrics under the settings of \tabref{tabrescomparison}.

\qual

In \tabref{tabrescomparisonNoPW}, all the methods do not use 3DPW for training. We use different settings for our model training.  Specifically, we use train sets of MPI-INF-3DHP, Human3.6M, InstaVariety and PoseTrack for training models evaluating on 3DPW and MPI-INF-3DHP test sets. The model evaluating on Human3.6M test set is trained with MPI-INF-3DHP, Human3.6M and InstaVariety train sets. Both 3DPW with true SMPL parameter labels and Human3.6M with pseudo SMPL parameter labels from Mosh \cite{loper2014mosh} are applied with SMPL loss. We compare our TePose with both frame-based models and video-based models. As shown in the table, our method provides competitive results comparing with previous methods. The acceleration error is high for frame-based methods, which is reasonable as they do not use temporal information. For all the three test sets, our model has better performance than other methods on PA-MPJPE and Accel. For the other metrics, our model also provides close performance compared with state-of-the-art. And for all the other video-based models, they need much longer videos as input, which means more waiting time and more complex computation. The quantitative results show that proposed system outperforms the previous state-of-the-art methods in both per-frame prediction accuracy and temporal consistency by introducing temporally embedded encoder and GCN-based adversarial training.

To further qualitatively demonstrate the improvement of proposed method, we compare the pose and shape estimation of TePose with VIBE \cite{kocabas2020vibe} and TCMR \cite{choi2021beyond} in \figref{qual}. Three in-the-wild samples from 3DPW test set and three from MPI-INF-3DHP test set are selected to show the difference. The results show that the inferences of proposed TePose are more accurate and stable than VIBE and TCMR. Both quantitative and qualitative results demonstrate the effectiveness of proposed method.

\tabablation

\subsection{Ablation Study}
In this section, we show how each module of our proposed system contributes to the improvement of human body pose and shape estimation performance. In \tabref{tabablation}, we show the performance of 4 different module combinations. These models are all trained on MPI-INF-3DHP, Human3.6M and InstaVariety while all of them are evaluated on the test set of Human3.6M, which has the same setting with the last three columns in \tabref{tabrescomparisonNoPW}. The first row in \tabref{tabablation} removes all the proposed modules. It uses single uni-directional GRU as temporal encoder. Compared with VIBE \cite{kocabas2020vibe}, the number of frames for model input is 6 instead of 16 in VIBE, and the adversarial learning of VIBE is also removed. It is reasonable to have worse performance than both VIBE and proposed TePose. The second row utilizes two GRUs as temporal encoder. One GRU is uni-directional and another one is bi-directional. It improves prediction performance compared with single-GRU model. When we add the temporally embedded encoder in the third row, SMPL parameter predictions for previous frames are needed as input for current-frame pose and shape estimation, which means the sequential data loading strategy also has to be added to make the training feasible. By adding the temporally embedded encoder and sequential data loading strategy in the third row of \tabref{tabablation}, we demonstrate the effectiveness of these two proposed modules. In the last row of this table, we further add the adversarial learning module using proposed GCN-based motion discriminator to make it become the complete TePose. The results also demonstrate it can improve the accuracy of pose prediction. In conclusion, the results from different combinations of proposed modules demonstrate the effectiveness for each of them.

\section{Conclusion}

We present TePose to estimate 3D human body pose and shape for live stream and video data. TePose utilizes mesh parameter predictions from previous frames as bridge to bring temporal information into prediction network for improving the accuracy and temporal consistency of overall estimation. Graph convolutional network is used to distinguish estimated pose sequences from real ones in dataset, so as to introduce extra adversarial loss for data samples without 3D supervision. A sequential data loading strategy is presented to make the convergence of the model using mixture input from previous mesh predictions and groundtruth mesh parameters. Compared to the previous methods, our approach provides competitive performance on accuracy and temporal consistency. By using temporal embedded input, TePose  leads to better human pose and body estimation with less previous frames as input. It will be helpful especially for live stream scenarios.

{\small
\bibliographystyle{ieee_fullname}
\bibliography{ref}
}

\end{document}